\documentclass{article}

\usepackage[dblblindworkshop, final]{neurips_2025}

\usepackage[utf8]{inputenc} 
\usepackage[T1]{fontenc}    
\usepackage{hyperref}       
\usepackage{url}            
\usepackage{booktabs}       
\usepackage{amsfonts}       
\usepackage{nicefrac}       
\usepackage{microtype}      
\usepackage{xcolor}         
\usepackage[most]{tcolorbox} 
\tcbuselibrary{listings,skins,breakable}
\usepackage{lmodern} 
\usepackage[T1]{fontenc}

\usepackage{multirow}
\usepackage{tcolorbox}
\usepackage{tabularx}

\PassOptionsToPackage{table}{xcolor}

\usepackage{wrapfig}
\usepackage{graphicx}

\usepackage{color, colortbl}
\usepackage[absolute,overlay]{textpos}

\definecolor{darkblue}{rgb}{0, 0, 0.5}
\hypersetup{colorlinks=true, citecolor=darkblue, linkcolor=darkblue, urlcolor=darkblue}
\definecolor{Highlight}{rgb}{0.92,0.94,1}

\title{Learning How to Use Tools, Not Just When: Pattern-Aware Tool-Integrated Reasoning}

\author{
Ningning Xu$^{1}$\thanks{Equal contribution.} \quad
Yuxuan Jiang$^{2}$\footnotemark[1] \quad
Shubhashis Roy Dipta$^{2}$ \quad
Hengyuan Zhang$^{3}$\\
$^{1}$University of Georgia\\
$^{2}$University of Maryland, Baltimore County \\
$^{3}$The University of Hong Kong\\
\texttt{nx32943@uga.edu}
}

\begin{document}

\maketitle

\begin{abstract}
Tool-integrated reasoning (TIR) has become a key approach for improving large reasoning models (LRMs) on complex problems. Prior work has mainly studied when to invoke tools, while overlooking how tools are applied. We identify two common patterns: a \emph{calculator-pattern} that uses code for direct computation, and an \emph{algorithmic-pattern} that encodes problems as programs. Misaligned choices often cause failures even when reasoning is sound. We propose a two-stage framework that first builds code competence from both patterns and then aligns pattern selection with teacher preferences. Across challenging math datasets, our pattern-aware method substantially improves both code usage and accuracy—for instance, raising Code@1 on MATH500 from 64.0\% to 70.5\% and on AIME24 from 26.7\% to 50.0\%. These gains highlight the effectiveness of a pattern-aware approach for tool-integrated reasoning.

\end{abstract}

\section{Introduction}


Tool-integrated reasoning (TIR) has become a powerful paradigm for enhancing large reasoning models (LRMs)~\cite{wei2025autotir,chen2025r1,yu2025chain}. By interacting with external verifiers such as code interpreters, TIR enables models to produce executable reasoning steps, making answers more accurate, faithful, and verifiable~\cite{zheng2025deepresearcher}. Recent progress has shown gains from better timing of tool invocation, multi-round tool calls, and tighter integration of external computation~\cite{dong2025tool,wei2025autotir}.

While effective, a critical question remains: \textbf{can models reliably decide how to use tools once invoked?} We provide an initial exploration of this problem. Models often misalign their tool-use strategies with problem demands, applying code in a mechanical and context-insensitive way that prevents the tool from working effectively. As Figure~\ref{fig:tir} illustrates, on the factorial problem \(1000! \div (800! \times 2!)\) the model used the \emph{calculator-pattern} and failed due to overflow, while on the “first 10-digit prime in \(\pi\)” problem it used the \emph{algorithmic-pattern} and stalled without output. \emph{In both cases, switching to the complementary pattern succeeds}, indicating that the failures stem from \emph{pattern selection} rather than a lack of reasoning ability.

To address this challenge, we propose a two-stage learning framework. In the first stage, \textit{code competence acquisition}, the model learns to generate reliable code from both calculator- and algorithmic-pattern data, without teacher preferences, ensuring stable tool use. In the second stage, \textit{pattern preference alignment}, teacher signals guide the model to choose the more suitable pattern for each problem. Together, these stages equip LRMs to not only invoke tools effectively but also adapt their strategy of use.

\begin{figure}[t]
  \centering

  \includegraphics[width=1\linewidth,page=1]{intro_final.pdf}
\caption{\textbf{Pattern mismatch in tool-integrated reasoning.}
Left: in \(1000! \div (800! \times 2!)\), an algebraic approach succeeds while direct computation overflows.
Right: in finding the first 10-digit prime in \(\pi\), a symbolic approach fails from context limits while a scanning approach succeeds.
These cases show that success depends on the chosen tool-use pattern.}

  \label{fig:tir}
\end{figure}

Our experiments on challenging math benchmarks show substantial and consistent gains. 
The model’s ability to produce executable code (@1) increases from 64.0\% to 70.5\% on MATH500, indicating that \textit{code competence acquisition} markedly stabilizes tool use. 
At the same time, overall problem-solving accuracy rises from 26.7\% to 50.0\% on AIME24, demonstrating that \textit{pattern preference alignment} further improves strategy selection and reduces failures due to mismatched tool use. 
Together, these results validate that modeling \emph{how} tools are used—beyond merely invoking them—yields large improvements in both code reliability and end-task performance.  

This work makes three contributions.  
(1) We identify and formalize \emph{pattern mismatch} in TIR, showing that misaligned tool-use strategies can derail otherwise sound reasoning.  
(2) We propose a two-stage framework that first builds code competence from both calculator- and algorithmic-pattern data (without teacher preferences) and then aligns pattern choice with teacher signals.  
(3) We provide empirical evidence on MATH500 and AIME24 that this approach substantially improves code@1 (64.0\% $\rightarrow$ 70.5\%) and accuracy (26.7\% $\rightarrow$ 50.0\%), and we illustrate how switching patterns flips failure cases into successes (Fig.~\ref{fig:tir}).


\section{Related Works}
\subsection{Reasoning with LRMs}
Among various LLM abilities, foundational research in semantic matching~\cite{xue2023dual, xue2025structcoh} and temporal QA~\cite{xue2024qa} has paved the way for complex understanding. While domain adaptation~\cite{huang2023dtbs} and test-time integration~\cite{huang2025cosmic} enhance robustness, reasoning ability achieved by post-training~\cite{huang2026rlvr, lan2025mappo,jiang2026cornerstonesstumblingblocksdeciphering} stands as a key frontier. Given that traditional SFT may suffer from incomplete learning~\cite{xue2026sft}, recent works introduce uncertainty-aware reward modeling~\cite{xue2026reward}, dense feedback from LM critics~\cite{cao2024enhancing}, and preference-driven code generation~\cite{li2025preference} to refine alignment. Moreover, recent studies have advanced role-playing reasoning through human-actor emulation~\cite{chen2026actormindemulatinghumanactor} and improved continual learning via prototype-conditioned generative replay~\cite{chen-zeng-2025-prototype}.

The scope of reasoning has expanded into multimodal domains, where models now "think with 3D"~\cite{chen2025think} and utilize reinforced visual perception~\cite{chen2025visrl, chen2025sifthinker}. Advanced frameworks like OmniVideo-R1~\cite{chen2026omnivideo} and DV-Matcher~\cite{chen2025dv} further integrate audio-visual and geometric cues. Such multimodal intelligence significantly boosts retrieval performance, including chat-driven search~\cite{xie2025chat, xie2026conquer} and vision-driven video representation~\cite{xie2026hvd, xie2026delving}. Furthermore, safeguarding techniques~\cite{yu2024robust, lan2025contextual} and hallucination mitigation~\cite{yu2024mechanistic, ji2025calibrating} ensure the reliability of these complex systems.

To support deployment, system efficiency and continual learning have become indispensable. Inference is optimized via multi-tier KV cache management~\cite{chu2025mcam}, selective sharing~\cite{chu2025selective}, and adaptive MoE scheduling~\cite{chu2025dynamic, shen2025expertflow}. Moreover, lightweight techniques like parameter-efficient sampling~\cite{yao2024swift}, structured pruning~\cite{li2026sepprune}, and distillation~\cite{li2025frequency} enable efficient task execution, from medical imaging~\cite{Li2025Efficient} to web agents~\cite{ding2026dynaweb}. Finally, reversible alignment~\cite{xiao2026reversible} and unified LoRA generation~\cite{xiaometa, chen2024three} allow models to learn continually, while comprehensive interaction surveys~\cite{li2026comprehensive} and robust federated learning~\cite{chen2024confusion} continue to push the boundaries of scalable AI.

Among LLMs' various abilities ~\cite{xue2026resilientroutingriskawaredynamic, zhao2026nonintrusivegraphbasedbotdetection, qi2026detectingabnormaluserfeedback}, subsequent work has focused on improving the quality of these reasoning paths, either by introducing diversity through self-consistency~\cite{wang2022self}, refining intermediate steps via process supervision~\cite{lightman2023let,jiang2025drp}, or encouraging verification of intermediate results~\cite{zelikman2022star, uesato2022solving,zhang2025find}. 

Beyond core architecture, recent research emphasizes a data-centric perspective on the entire LLM lifecycle~\cite{rao2025data}, where mathematical reasoning is further enhanced by adversarial data synthesis~\cite{yu2026mathagent} and iterative DPO with dynamic sampling~\cite{rao2025dynamic}. To improve model robustness and learning efficiency, advanced techniques such as difficulty-based sample weighting~\cite{zhou2022understanding}, implicit data augmentation~\cite{zhou2024boosting}, and adversarial training with anti-adversaries~\cite{zhou2024adversarial} have been introduced.

In the context of embodied AI and spatio-temporal reasoning, unified token spaces for human-object interaction~\cite{yang2026unihoi} and object-centric decoupling for robotic manipulation~\cite{yang2026instrucrobo} have improved explainable decision-making. These abilities are further integrated into 3D scene understanding~\cite{yang2026unibvr} and autonomous driving through spatio-temporal Chain-of-Thought (CoT)~\cite{zeng2025fsdrive} and dual-memory vision-language navigation~\cite{zeng2025janusvln}. Furthermore, the application of multi-scale graph learning and Transformers has shown significant potential in predicting complex real-world dynamics, including financial fraud detection~\cite{lei2026multiscale}, traffic flow prediction~\cite{shen2026mftformer}, disaster risk assessment~\cite{shen2025aienhanced}, mental health~\cite{yang2026exploring}, recommendations~\cite{wang2025improving} and socio-economic metrics such as labor market gradients~\cite{liu2026health} and business resilience indices~\cite{Shen2026bri}.
\subsection{Tool-Integrated Reasoning}

Tool-integrated reasoning has emerged as a powerful paradigm for enhancing large reasoning models (LRMs)~\cite{wei2025autotir,chen2025r1}, enabling executable reasoning steps through external tools such as Python interpreters~\cite{zheng2025deepresearcher,jiang2026scribe}. Early work like PoT~\cite{chen2022program}, PAL~\cite{gao2023pal}, and MathPrompter~\cite{imani2023mathprompter} showed that converting reasoning into code execution or lightweight snippets can substantially improve math performance, while implementations such as ToRA~\cite{gou2023tora} and Qwen2.5-Math-Instruct-TIR~\cite{shao2024deepseekmath} further demonstrated gains from deeper integration. However, prior studies primarily focus on \emph{when} to call tools, overlooking \emph{how} tools are applied once invoked. We address this gap by modeling pattern selection in TIR, introducing a pattern-aware framework that aligns tool-use strategies and yields significant improvements in reasoning accuracy.

\begin{wrapfigure}{r}{0.5\textwidth}
    \centering
    \includegraphics[width=0.41\textwidth,height=0.30\textheight]{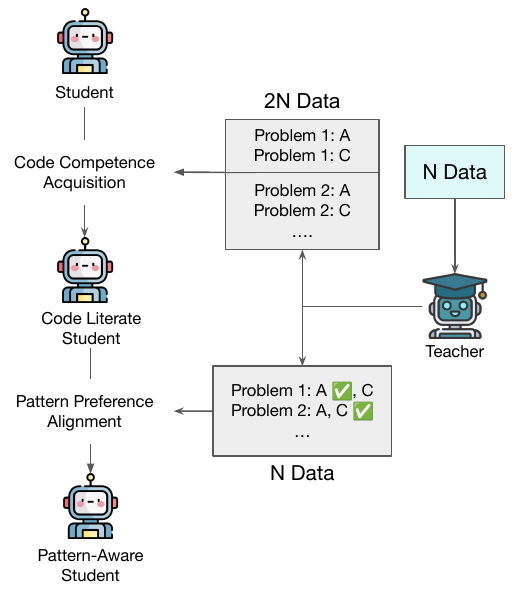}
    \vspace{-2mm}
    \caption{\textbf{Two-stage training framework.} 
In the 1\textsuperscript{st} phase, each problem is expanded into both calculator- and algorithmic-style solutions ($2N$ data) to build code competence, 
and in the 2\textsuperscript{nd} phase, teacher-provided preferences on $N$ problems guide the student to become pattern-aware.}
    \vspace{-15mm}
    \label{fig:method}
\end{wrapfigure}

\section{Method}
We train a pattern-aware reasoning model in two stages. For stage 1, we builds code competence by supervising the model on both calculator-style  and algorithmic-style solutions. Stage 2 performs pattern preference alignment, using Direct preference optimization (DPO) to learn the pattern preference.

\subsection{Training set construction}

We distinguish two usage patterns in tool-integrated reasoning: the calculator-pattern, where code is used only for direct computation or verification, and the algorithmic-pattern, where problems are expressed as full programs. For each problem, we generate a Chosen Solution with the more suitable pattern and a Counter Solution with the alternative one. This dual construction not only equips the model with competence in both styles of code usage, but also provides the supervision necessary to align its pattern selection with teacher preferences, thereby improving robustness and accuracy in downstream reasoning tasks. And We put the complete version is provided in Section~\ref{prompt}.




\subsection{Code Competence Acquisition}
For each problem $x$, we construct two solutions:
\begin{itemize}
    \item an algorithmic-style solution $y^{alg}$ that converts the problem to an execuable program;
    \item a calculator-style solution $y_{calc}$ that keeps the reasoing path while using python for arithmetic or verification.
\end{itemize}

We write both solutions with the same ouput shcema (reasoning $\rightarrow$ python block $\rightarrow$ ouputs $\rightarrow$ final answer), which can simplifies parsing and execution. Let $D_{SFT}=\{(x_i, y_i^{alg}), (x_i, y_i^{calc})\}$ be the set of both patterns, we minimize the negative log-likelihood over response tokens:
\[
\mathcal{L}_{\text{SFT}}
= -\mathbb{E}_{(x,y)\sim\mathcal{D}_{\text{SFT}}}
\sum_{t=1}^{|y|}\log \pi_\theta(y_t \mid x, y_{<t}),
\]

\subsection{Pattern Preference Alignment}
We use DPO~\cite{rafailov2023direct} to empower our model with the ability to pick the correct pattern to solve math problems. For each problem $x$, we collect two candidate completions $y^{alg}, y^{calc}$ and label a winner $y^{+}$ and loser $y^{-}$ using the teacher model. For a pair $(x, y^{+}, y^{-})$, we then apply direction preference optimization with a frozen reference policy $\pi_{ref}$ (our SFT model):
\[
\mathcal{L}_{\text{DPO}}
= -\log \sigma \Big(
\beta\big[
\log \pi_\theta(y^{+}\!\mid x) - \log \pi_\theta(y^{-}\!\mid x)
- \big(\log \pi_{\text{ref}}(y^{+}\!\mid x) - \log \pi_{\text{ref}}(y^{-}\!\mid x)\big)
\big] \Big),
\]

where $\beta$ scales preference strength and $\sigma$ is the logistic function. This increases the likelihood of preferred (pattern-appropriate) solutions relative to dispreferred ones, without training a separate reward model.



\section{Experiments}
\subsection{Dataset}
We use google Gemini-2.5-flash-lite as our teacher model, using our designed two pattern prompt to solve the problems in OpenR1-Math-220k~\cite{openr1math220k} dataset. From this corpus, we select the first 10,000 problems and divide them into training and validation sets using a 9:1 ratio.
To assess complex mathematical reasoning and generalization, we evaluate on a broad set of out-of-domain benchmarks, including MATH500~\cite{math}, AIME24~\cite{aime2024}, and AMC23~\cite{amc2024}.

\subsection{Evaluation Metrics}
We report \textbf{Pass@1} as the primary accuracy metric. For additional insight into reasoning behavior, we also track (i) the proportion of problems where the model’s output contains executable code (\textbf{Code@1}), and (ii) the proportion of problems where such code is both present and leads to the correct final answer (\textbf{Code+Pass@1}). 

\section{Main Results}


\begin{table*}[htbp]
\centering
\small
\resizebox{0.95\textwidth}{!}{%
\begin{tabular}{l|cc|cc|cc|cc}
\toprule
\textbf{Method} 
& \multicolumn{2}{c|}{\textbf{R1Math}} 
& \multicolumn{2}{c|}{\textbf{GSM8K (OOD)}} 
& \multicolumn{2}{c|}{\textbf{MATH500 (OOD)}} 
& \multicolumn{2}{c}{\textbf{AIME24 (OOD)}} \\
& Code@1 & Code+Pass@1 
& Code@1 & Code+Pass@1
& Code@1 & Code+Pass@1
& Code@1 & Code+Pass@1 \\
\midrule



\multicolumn{9}{l}{\textbf{R1-Distill-Qwen-1.5B}} \\
\midrule
Base      & 3.0\%   & 2.0\%   & 4.2\%   & 4.2\%   & 8.0\%   & 6.0\%   & 0.0\%   & 0.0\% \\
+SFT      & 66.0\%  & 55.5\%  & 72.0\%  & 70.4\%  & 64.0\%  & 58.2\%  & 26.7\%  & 6.6\% \\

\rowcolor{Highlight}
\textbf{+SFT+DPO}  & \textbf{72.2\%} & \textbf{70.3\%}
                   & \textbf{80.1\%} & \textbf{77.4\%}
                   & \textbf{70.5\%} & \textbf{63.2\%}
                   & \textbf{50.0\%} & \textbf{26.7\%} \\

\bottomrule

\end{tabular}
} 
\caption{Pass@1 and Code+Pass@1 accuracy on R1-Distill-Qwen models across R1Math, GSM8K, MATH500, and AIME24.}
\label{main_results}
\end{table*}

According to Table~\ref{main_results}, the base model achieves near-zero code usage and accuracy. With SFT, the model begins to leverage Python extensively (e.g., R1Math Code@1 rises to 66\%), and correctness follows accordingly. Adding DPO yields further substantial improvements, notably pushing GSM8K Code@1 from 72.0\% to 80.1\% and AIME24 from 26.7\% to 50.0\%, while also boosting correctness (e.g., AIME24 Code+Pass@1 from 6.6\% to 26.7\%). These results highlight that SFT is crucial for enabling tool use, whereas DPO markedly enhances both adoption and accuracy.

\section{Conclusion}

In this work, we studied two complementary tool-use patterns—algorithmic code generation and calculator-style usage—and showed that many failures arise from mismatched pattern choices rather than reasoning ability. Experiments on GSM8K, MATH500, and AIME24 demonstrate that our pattern-aware approach improves strategy selection and accuracy, offering an effective and lightweight way to enhance tool-integrated reasoning.

\medskip
{
\small
\bibliographystyle{plain} 
\bibliography{ref}

@article{yu2025chain,
  title={Chain-of-Reasoning: Towards Unified Mathematical Reasoning in Large Language Models via a Multi-Paradigm Perspective},
  author={Yu, Yiyao and Zhang, Yuxiang and Zhang, Dongdong and Liang, Xiao and Zhang, Hengyuan and Zhang, Xingxing and Yang, Ziyi and Khademi, Mahmoud and Awadalla, Hany and Wang, Junjie and others},
  journal={arXiv preprint arXiv:2501.11110},
  year={2025}
}

@article{rafailov2023direct,
  title={Direct preference optimization: Your language model is secretly a reward model},
  author={Rafailov, Rafael and Sharma, Archit and Mitchell, Eric and Manning, Christopher D and Ermon, Stefano and Finn, Chelsea},
  journal={Advances in neural information processing systems},
  volume={36},
  pages={53728--53741},
  year={2023}
}

@misc{openr1math220k,
  title        = {OpenR1-Math-220k},
  author       = {{OpenR1 Team}},
  year         = {2025},
  howpublished = {\url{https://huggingface.co/datasets/open-r1/OpenR1-Math-220k}},
  note         = {Accessed: 2025-09-25}
}

@article{math,
  title={Measuring mathematical problem solving with the math dataset},
  author={Hendrycks, Dan and Burns, Collin and Kadavath, Saurav and Arora, Akul and Basart, Steven and Tang, Eric and Song, Dawn and Steinhardt, Jacob},
  journal={arXiv preprint arXiv:2103.03874},
  year={2021}
}

@misc{aime2024,
  author       = {{AI-MO Team}},
  title        = {{AIMO Validation Set - AIME Subset}},
  howpublished = {\url{https://huggingface.co/datasets/AI-MO/aimo-validation-aime}},

  year         = {2024}
}

@misc{amc2024,
  author       = {{AI-MO Team}},
  title        = {{AIMO Validation Set - AMC Subset}},
  howpublished = {\url{https://huggingface.co/datasets/AI-MO/aimo-validation-amc}},

  year         = {2024}
}

@article{wang2022self,
  title={Self-consistency improves chain of thought reasoning in language models},
  author={Wang, Xuezhi and Wei, Jason and Schuurmans, Dale and Le, Quoc and Chi, Ed and Narang, Sharan and Chowdhery, Aakanksha and Zhou, Denny},
  journal={arXiv preprint arXiv:2203.11171},
  year={2022}
}

@inproceedings{lightman2023let,
  title={Let's verify step by step},
  author={Lightman, Hunter and Kosaraju, Vineet and Burda, Yuri and Edwards, Harrison and Baker, Bowen and Lee, Teddy and Leike, Jan and Schulman, John and Sutskever, Ilya and Cobbe, Karl},
  booktitle={The Twelfth International Conference on Learning Representations},
  year={2023}
}

@article{zelikman2022star,
  title={Star: Bootstrapping reasoning with reasoning},
  author={Zelikman, Eric and Wu, Yuhuai and Mu, Jesse and Goodman, Noah},
  journal={Advances in Neural Information Processing Systems},
  volume={35},
  pages={15476--15488},
  year={2022}
}

@article{uesato2022solving,
  title={Solving math word problems with process-and outcome-based feedback},
  author={Uesato, Jonathan and Kushman, Nate and Kumar, Ramana and Song, Francis and Siegel, Noah and Wang, Lisa and Creswell, Antonia and Irving, Geoffrey and Higgins, Irina},
  journal={arXiv preprint arXiv:2211.14275},
  year={2022}
}

@article{chen2022program,
  title={Program of thoughts prompting: Disentangling computation from reasoning for numerical reasoning tasks},
  author={Chen, Wenhu and Ma, Xueguang and Wang, Xinyi and Cohen, William W},
  journal={arXiv preprint arXiv:2211.12588},
  year={2022}
}

@inproceedings{gao2023pal,
  title={Pal: Program-aided language models},
  author={Gao, Luyu and Madaan, Aman and Zhou, Shuyan and Alon, Uri and Liu, Pengfei and Yang, Yiming and Callan, Jamie and Neubig, Graham},
  booktitle={International Conference on Machine Learning},
  pages={10764--10799},
  year={2023},
  organization={PMLR}
}

@article{imani2023mathprompter,
  title={Mathprompter: Mathematical reasoning using large language models},
  author={Imani, Shima and Du, Liang and Shrivastava, Harsh},
  journal={arXiv preprint arXiv:2303.05398},
  year={2023}
}

@article{gou2023tora,
  title={Tora: A tool-integrated reasoning agent for mathematical problem solving},
  author={Gou, Zhibin and Shao, Zhihong and Gong, Yeyun and Shen, Yelong and Yang, Yujiu and Huang, Minlie and Duan, Nan and Chen, Weizhu},
  journal={arXiv preprint arXiv:2309.17452},
  year={2023}
}

@article{shao2024deepseekmath,
  title={Deepseekmath: Pushing the limits of mathematical reasoning in open language models},
  author={Shao, Zhihong and Wang, Peiyi and Zhu, Qihao and Xu, Runxin and Song, Junxiao and Bi, Xiao and Zhang, Haowei and Zhang, Mingchuan and Li, YK and Wu, Yang and others},
  journal={arXiv preprint arXiv:2402.03300},
  year={2024}
}

@article{jiang2025drp,
  title={DRP: Distilled Reasoning Pruning with Skill-aware Step Decomposition for Efficient Large Reasoning Models},
  author={Jiang, Yuxuan and Li, Dawei and Ferraro, Frank},
  journal={arXiv preprint arXiv:2505.13975},
  year={2025}
}

@article{li2025generation,
  title={From generation to judgment: Opportunities and challenges of llm-as-a-judge, 2025},
  author={Li, Dawei and Jiang, Bohan and Huang, Liangjie and Beigi, Alimohammad and Zhao, Chengshuai and Tan, Zhen and Bhattacharjee, Amrita and Jiang, Yuxuan and Chen, Canyu and Wu, Tianhao and others},
  journal={URL https://arxiv. org/abs/2411.16594},
  year={2025}
}

@article{wei2025autotir,
  title={Autotir: Autonomous tools integrated reasoning via reinforcement learning},
  author={Wei, Yifan and Yu, Xiaoyan and Weng, Yixuan and Pan, Tengfei and Li, Angsheng and Du, Li},
  journal={arXiv preprint arXiv:2507.21836},
  year={2025}
}

@article{chen2025r1,
  title={R1-Code-Interpreter: Training LLMs to Reason with Code via Supervised and Reinforcement Learning},
  author={Chen, Yongchao and Liu, Yueying and Zhou, Junwei and Hao, Yilun and Wang, Jingquan and Zhang, Yang and Fan, Chuchu},
  journal={arXiv preprint arXiv:2505.21668},
  year={2025}
}

@article{zheng2025deepresearcher,
  title={Deepresearcher: Scaling deep research via reinforcement learning in real-world environments},
  author={Zheng, Yuxiang and Fu, Dayuan and Hu, Xiangkun and Cai, Xiaojie and Ye, Lyumanshan and Lu, Pengrui and Liu, Pengfei},
  journal={arXiv preprint arXiv:2504.03160},
  year={2025}
}

@article{dong2025tool,
  title={Tool-Star: Empowering LLM-Brained Multi-Tool Reasoner via Reinforcement Learning},
  author={Dong, Guanting and Chen, Yifei and Li, Xiaoxi and Jin, Jiajie and Qian, Hongjin and Zhu, Yutao and Mao, Hangyu and Zhou, Guorui and Dou, Zhicheng and Wen, Ji-Rong},
  journal={arXiv preprint arXiv:2505.16410},
  year={2025}
}

@article{zhang2025find,
  title={Find Your Optimal Teacher: Personalized Data Synthesis via Router-Guided Multi-Teacher Distillation},
  author={Zhang, Hengyuan and Yang, Shiping and Liang, Xiao and Shang, Chenming and Jiang, Yuxuan and Tao, Chaofan and Xiong, Jing and So, Hayden Kwok-Hay and Xie, Ruobing and Chang, Angel X and others},
  journal={arXiv preprint arXiv:2510.10925},
  year={2025}
}

@article{gao2023human,
  title={Human-like summarization evaluation with chatgpt},
  author={Gao, Mingqi and Ruan, Jie and Sun, Renliang and Yin, Xunjian and Yang, Shiping and Wan, Xiaojun},
  journal={arXiv preprint arXiv:2304.02554},
  year={2023}
}

@article{jiang2026scribe,
  title={SCRIBE: Structured Mid-Level Supervision for Tool-Using Language Models},
  author={Jiang, Yuxuan and Ferraro, Francis},
  journal={arXiv preprint arXiv:2601.03555},
  year={2026}
}

@misc{xue2026resilientroutingriskawaredynamic,

      title={Resilient Routing: Risk-Aware Dynamic Routing in Smart Logistics via Spatiotemporal Graph Learning}, 

      author={Zhiming Xue and Sichen Zhao and Yalun Qi and Xianling Zeng and Zihan Yu},

      year={2026},

      eprint={2601.13632},

      archivePrefix={arXiv},

      primaryClass={cs.AI},

      url={https://arxiv.org/abs/2601.13632}, 

}

@misc{zhao2026nonintrusivegraphbasedbotdetection,

      title={Non-Intrusive Graph-Based Bot Detection for E-Commerce Using Inductive Graph Neural Networks}, 

      author={Sichen Zhao and Zhiming Xue and Yalun Qi and Xianling Zeng and Zihan Yu},

      year={2026},

      eprint={2601.22579},

      archivePrefix={arXiv},

      primaryClass={cs.LG},

      url={https://arxiv.org/abs/2601.22579}, 

}

@misc{qi2026detectingabnormaluserfeedback,

      title={Detecting Abnormal User Feedback Patterns through Temporal Sentiment Aggregation}, 

      author={Yalun Qi and Sichen Zhao and Zhiming Xue and Xianling Zeng and Zihan Yu},

      year={2026},

      eprint={2604.00020},

      archivePrefix={arXiv},

      primaryClass={cs.CL},

      url={https://arxiv.org/abs/2604.00020}, 

}

@article{huang2026rlvr,
  title={Semantic-Space Exploration and Exploitation in RLVR for LLM Reasoning},
  author={Huang, Fanding and Huang, Guanbo and Fan, Xiao and He, Yi and Liang, Xiao and Chen, Xiao and Jiang, Qinting and Khan, Faisal Nadeem and Jiang, Jingyan and Wang, Zhi},
  journal={arXiv preprint arXiv:2509.23808},
  year={2026}
}

@inproceedings{huang2025cosmic,
  title={COSMIC: Clique-Oriented Semantic Multi-space Integration for Robust CLIP Test-Time Adaptation},
  author={Huang, Fanding and Jiang, Jingyan and Jiang, Qinting and Li, Hebei and Khan, Faisal Nadeem and Wang, Zhi},
  booktitle={CVPR},
  pages={9772--9781},
  year={2025}
}

@inproceedings{huang2023dtbs,
  title={DTBS: Dual-Teacher Bi-Directional Self-Training for Domain Adaptation in Nighttime Semantic Segmentation},
  author={Huang, Fanding and Yao, Zihao and Zhou, Wenhui},
  booktitle={ECAI},
  pages={1084--1091},
  year={2023}
}

@inproceedings{xue2024qa,
  title={Question Calibration and Multi-hop Modeling for Temporal Question Answering},
  author={Xue, Chao and Liang, Di and Wang, Pengfei and Zhang, Jing},
  booktitle={AAAI},
  volume={38},
  number={17},
  pages={19332--19340},
  year={2024}
}

@inproceedings{xue2023dual,
  title={Dual Path Modeling for Semantic Matching by Perceiving Subtle Conflicts},
  author={Xue, Chao and Liang, Di and Wang, Sirui and Zhang, Jing and Wu, Wei},
  booktitle={ICASSP},
  pages={1--5},
  year={2023}
}

@inproceedings{xue2025structcoh,
  title={StructCoh: Structured Contrastive Learning for Context-Aware Text Semantic Matching},
  author={Xue, Chao and Gao, Ziyuan},
  booktitle={PRICAI},
  pages={300--315},
  year={2025}
}

@misc{jiang2026cornerstonesstumblingblocksdeciphering,
      title={Cornerstones or Stumbling Blocks? Deciphering the Rock Tokens in On-Policy Distillation}, 
      author={Yuxuan Jiang and Runchao Li and Shubhashis Roy Dipta and Dawei Li and Zhao Yang},
      year={2026},
      eprint={2605.09253},
      archivePrefix={arXiv},
      primaryClass={cs.CL},
      url={https://arxiv.org/abs/2605.09253}, 
}

@misc{xue2026sft,
  title={Why Supervised Fine-Tuning Fails to Learn: A Systematic Study of Incomplete Learning in Large Language Models}, 
  author={Xue, Chao and others},
  year={2026},
  eprint={2604.10079}
}

@misc{xue2026reward,
  title={Reason Only When Needed: Efficient Generative Reward Modeling via Model-Internal Uncertainty}, 
  author={Xue, Chao and others},
  year={2026},
  eprint={2604.10072}
}

@article{lan2025mappo,
  title={MaPPO: Maximum a Posteriori Preference Optimization with Prior Knowledge},
  author={Lan, Guangchen and others},
  journal={arXiv preprint arXiv:2507.21183},
  year={2025}
}

@inproceedings{lan2025contextual,
  title={Contextual Integrity in LLMs via Reasoning and Reinforcement Learning},
  author={Lan, Guangchen and others},
  booktitle={NeurIPS},
  year={2025}
}

@article{li2025preference,
  title={A preference-driven methodology for efficient code generation},
  author={Li, Yuqi and others},
  journal={IEEE Transactions on Artificial Intelligence},
  year={2025}
}

@inproceedings{cao2024enhancing,
  title={Enhancing reinforcement learning with dense rewards from language model critic},
  author={Cao, Meng and others},
  booktitle={EMNLP},
  year={2024}
}

@article{chen2025think,
  title={Think with 3D: Geometric Imagination Grounded Spatial Reasoning from Limited Views},
  author={Chen, Zhangquan and others},
  journal={arXiv preprint arXiv:2510.18632},
  year={2025}
}

@article{chen2025visrl,
  title={Visrl: Intention-driven visual perception via reinforced reasoning},
  author={Chen, Zhangquan and others},
  journal={arXiv preprint arXiv:2503.07523},
  year={2025}
}

@article{chen2026omnivideo,
  title={OmniVideo-R1: Reinforcing Audio-visual Reasoning with Query Intention and Modality Attention},
  author={Chen, Zhangquan and others},
  journal={arXiv preprint arXiv:2602.05847},
  year={2026}
}

@article{chen2025sifthinker,
  title={SIFThinker: Spatially-Aware Image Focus for Visual Reasoning},
  author={Chen, Zhangquan and others},
  journal={arXiv preprint arXiv:2508.06259},
  year={2025}
}

@inproceedings{chen2025dv,
  title={DV-Matcher: Deformation-based Non-Rigid Point Cloud Matching Guided by Pre-trained Visual Features},
  author={Chen, Zhangquan and others},
  booktitle={CVPR},
  year={2025}
}

@inproceedings{xie2025chat,
  title={Chat-driven text generation and interaction for person retrieval},
  author={Xie, Zequn and others},
  booktitle={EMNLP},
  year={2025}
}

@article{xie2026hvd,
  title={HVD: Human Vision-Driven Video Representation Learning for Text-Video Retrieval},
  author={Xie, Zequn and others},
  journal={arXiv preprint arXiv:2601.16155},
  year={2026}
}

@article{xie2026conquer,
  title={CONQUER: Context-Aware Representation with Query Enhancement for Text-Based Person Search},
  author={Xie, Zequn},
  journal={arXiv preprint arXiv:2601.18625},
  year={2026}
}

@article{xie2026delving,
  title={Delving deeper: Hierarchical visual perception for robust video-text retrieval},
  author={Xie, Zequn and others},
  journal={arXiv preprint arXiv:2601.12768},
  year={2026}
}

@article{chu2025selective,
  title={Selective kv-cache sharing to mitigate timing side-channels in llm inference},
  author={Chu, Kexin and others},
  journal={arXiv preprint arXiv:2508.08438},
  year={2025}
}

@inproceedings{chu2025mcam,
  title={MCaM: Efficient LLM Inference with Multi-tier KV Cache Management},
  author={Chu, Kexin and others},
  booktitle={ICDCS},
  year={2025}
}

@article{chu2025dynamic,
  title={Dynamic Expert Quantization for Scalable Mixture-of-Experts Inference},
  author={Chu, Kexin and others},
  journal={arXiv preprint arXiv:2511.15015},
  year={2025}
}

@article{shen2025expertflow,
  title={ExpertFlow: Adaptive Expert Scheduling and Memory Coordination for Efficient MoE Inference},
  author={Shen, Zixu and others},
  journal={arXiv preprint arXiv:2510.26730},
  year={2025}
}

@article{yao2024swift,
  title={Swift sampler: Efficient learning of sampler by 10 parameters},
  author={Yao, Jiawei and others},
  journal={NeurIPS},
  year={2024}
}

@article{chen2024confusion,
  title={Confusion-resistant federated learning via diffusion-based data harmonization on non-IID data},
  author={Chen, Xiaohong and others},
  journal={NeurIPS},
  year={2024}
}

@inproceedings{xiao2026reversible,
  title={Reversible primitive--composition alignment for continual vision--language learning},
  author={Xiao, Canran and others},
  booktitle={ICLR},
  year={2026}
}

@inproceedings{xiaometa,
  title={Meta-UCF: Unified Task-Conditioned LoRA Generation for Continual Learning in Large Language Models},
  author={Xiao, ShiLin and others},
  booktitle={ICLR},
  year={2026}
}

@inproceedings{li2025frequency,
  title={Frequency-aligned knowledge distillation for lightweight spatiotemporal forecasting},
  author={Li, Yuqi and others},
  booktitle={ICCV},
  year={2025}
}

@inproceedings{li2026sepprune,
  title={Sepprune: Structured pruning for efficient deep speech separation},
  author={Li, Yuqi and others},
  booktitle={AAAI},
  year={2026}
}

@article{li2026comprehensive,
  title={A Comprehensive Survey of Interaction Techniques in 3D Scene Generation},
  author={Li, Yuqi and others},
  journal={Authorea},
  year={2026}
}

@article{Li2025Efficient,
  title={Efficient Medical Image Segmentation via Reinforcement Learning-Driven K-Space Sampling},
  author={Li, Yuqi and others},
  journal={IEEE TETCI},
  year={2025}
}

@article{yu2024robust,
  title={Robust LLM safeguarding via refusal feature adversarial training},
  author={Yu, Lei and others},
  journal={arXiv preprint arXiv:2409.20089},
  year={2024}
}

@inproceedings{yu2024mechanistic,
  title={Mechanistic understanding and mitigation of language model non-factual hallucinations},
  author={Yu, Lei and others},
  booktitle={EMNLP},
  year={2024}
}

@inproceedings{ji2025calibrating,
  title={Calibrating Verbal Uncertainty as a Linear Feature to Reduce Hallucinations},
  author={Ji, Ziwei and others},
  booktitle={EMNLP},
  year={2025}
}

@article{ding2026dynaweb,
  title={DynaWeb: Model-Based Reinforcement Learning of Web Agents},
  author={Ding, Hang and others},
  journal={arXiv preprint arXiv:2601.22149},
  year={2026}
}

@inproceedings{chen2024three,
  title={A Three-Phases-LORA Finetuned Hybrid LLM Integrated with Strong Prior Module in the Education Context},
  author={Chen, Zhangquan and others},
  booktitle={ICANN},
  year={2024}
}

@misc{chen2026actormindemulatinghumanactor,
      title={ActorMind: Emulating Human Actor Reasoning for Speech Role-Playing}, 
      author={Xi Chen and Wei Xue and Yike Guo},
      year={2026},
      eprint={2604.11103},
      archivePrefix={arXiv},
      primaryClass={cs.SD},
      url={https://arxiv.org/abs/2604.11103}, 
}

@misc{yu2026mathagent,
  title={MathAgent: Adversarial Evolution of Constraint Graphs for Mathematical Reasoning Data Synthesis}, 
  author={Yu, Zixiong and others},
  year={2026},
  eprint={2604.11188},
  archivePrefix={arXiv}
}

@misc{rao2025dynamic,
  title={Dynamic Sampling that Adapts: Iterative DPO for Self-Aware Mathematical Reasoning}, 
  author={Rao, Jun and others},
  year={2025},
  eprint={2505.16176},
  archivePrefix={arXiv}
}

@article{rao2025data,
  title={A Data-Centric Perspective on the Lifecycle of Large Language Models},
  author={Rao, Jun and others},
  journal={TechRxiv},
  year={2025},
  doi={10.36227/techrxiv.176620610.03288677/v1}
}

@inproceedings{yang2026unihoi,
  title={UniHOI: Unified Human-Object Interaction Understanding via Unified Token Space},
  author={Yang, Panqi and others},
  booktitle={AAAI},
  year={2026}
}

@article{yang2026instrucrobo,
  title={InstrucRobo: Object-centric multi-instruction decoupling model for explainable robotic manipulation},
  author={Yang, Panqi and others},
  journal={Engineering Applications of Artificial Intelligence},
  volume={171},
  year={2026}
}

@article{yang2026unibvr,
  title={UniBVR: Balancing visual and reasoning abilities in unified 3D scene understanding},
  author={Yang, Panqi and others},
  journal={Neurocomputing},
  volume={671},
  year={2026}
}

@article{zeng2025fsdrive,
  title={FutureSightDrive: Thinking Visually with Spatio-Temporal CoT for Autonomous Driving},
  author={Zeng, Shuang and others},
  journal={arXiv preprint arXiv:2505.17685},
  year={2025}
}

@article{zeng2025janusvln,
  title={JanusVLN: Decoupling Semantics and Spatiality with Dual Implicit Memory for Vision-Language Navigation},
  author={Zeng, Shuang and others},
  journal={arXiv preprint arXiv:2509.22548},
  year={2025}
}

@article{lei2026multiscale,
  title={A Multi-Scale Graph Learning Framework for Financial Fraud Detection},
  author={Lei, Yiming and others},
  year={2026},
  eprint={2603.14592},
  archivePrefix={arXiv}
}

@article{shen2026mftformer,
  title={MFTFormer: Meteorological-Frequency-Temporal Transformer for Traffic Flow Prediction},
  author={Shen, Qiannan and others},
  journal={Research Square},
  year={2026}
}

@article{shen2025aienhanced,
  title={AI-Enhanced Disaster Risk Prediction with Explainable SHAP Analysis},
  author={Shen, Qiannan and others},
  journal={Research Square},
  year={2025}
}

@article{liu2026health,
  title={The Health-Wealth Gradient in Labor Markets},
  author={Liu, Dingyuan and others},
  journal={Computation},
  year={2026}
}

@article{Shen2026bri,
  title={Business Resilience Index (BRI): Evaluating Economic Recovery},
  author={Shen, Qiannan and others},
  journal={Sustainability},
  year={2026}
}

@article{yang2026exploring,
  title={Exploring the application boundaries of llms in mental health: A systematic scoping review},
  author={Yang, Jinhua and Liu, Ting and Luo, Yiming Taclis and Niu, Tianyue and Pang, Patrick and Xiang, Ao and Yang, Qin},
  journal={Frontiers in Psychology},
  volume={16},
  pages={1715306},
  year={2026}
}

@inproceedings{wang2025improving,
  title={Improving Sequential Recommendations with TokenLevel LLM Representatio},
  author={Wang, Dingzhou and Chang, Lu and Men, Luyao and He, Jiajun and Yang, Yinuo and Liang, Yefeng},
  booktitle={2025 4th International Conference on Cloud Computing, Big Data Application and Software Engineering (CBASE)},
  pages={509--514},
  year={2025},
  organization={IEEE}
}

@article{zhou2024boosting,
  title={Boosting model resilience via implicit adversarial data augmentation},
  author={Zhou, Xiaoling and others},
  journal={arXiv preprint arXiv:2404.16307},
  year={2024}
}

@article{zhou2024adversarial,
  title={Adversarial training with anti-adversaries},
  author={Zhou, Xiaoling and others},
  journal={IEEE TPAMI},
  year={2024}
}

@inproceedings{zhou2022understanding,
  title={Understanding difficulty-based sample weighting},
  author={Zhou, Xiaoling and others},
  booktitle={ECML PKDD},
  year={2022}
}

@inproceedings{chen-zeng-2025-prototype,
    title = "Prototype Conditioned Generative Replay for Continual Learning in {NLP}",
    author = "Chen, Xi  and
      Zeng, Min",
    editor = "Chiruzzo, Luis  and
      Ritter, Alan  and
      Wang, Lu",
    booktitle = "Proceedings of the 2025 Conference of the Nations of the Americas Chapter of the Association for Computational Linguistics: Human Language Technologies (Volume 1: Long Papers)",
    month = apr,
    year = "2025",
    address = "Albuquerque, New Mexico",
    publisher = "Association for Computational Linguistics",
    url = "https://aclanthology.org/2025.naacl-long.636/",
    doi = "10.18653/v1/2025.naacl-long.636",
    pages = "12754--12770",
    ISBN = "979-8-89176-189-6",
    }

@article{jiang2024memorization,
  title={Memorization Over Reasoning? Exposing and Mitigating Verbatim Memorization in Large Language Models' Character Understanding Evaluation},
  author={Jiang, Yuxuan and Ferraro, Francis},
  journal={arXiv preprint arXiv:2412.14368},
  year={2024}
}
}


\appendix
\section{Complete Prompt}
\label{prompt}
\begin{tcolorbox}[colback=gray!5, colframe=gray!50, title=Double Pattern Prompt]
{\slshape 
Use a python interpreter as a tool to solve the following math problem.

There are two possible usage patterns:
\begin{itemize}
    \item Pattern A: Treat the problem as a coding problem, write a complete solution using Python code.
    \item Pattern B: Treat the Python interpreter as a simple calculator, only using it for arithmetic or verification when needed.
\end{itemize}

Your task:
\begin{enumerate}
    \item Decide which pattern (A or B) is more appropriate for this problem. Call this the \textbf{Chosen Solution}.
    \item Provide the \textbf{Chosen Solution} with EXACTLY ONE continuous reasoning paragraph (no lists, no numbering, no bullet points). Then give ONE Python code block, then the code outputs, then the final answer.
    \item Provide the \textbf{Counterfactual Solution} using the other pattern with the SAME constraints (one continuous reasoning paragraph, one code block, outputs, final answer).
    \item Strict constraints:
   - Do NOT restate the problem text.
   - Do NOT include introductions, apologies, or meta comments.
   - Do NOT duplicate any content across fields.
\end{enumerate}
Output format (pure JSON, no backticks, no extra keys):
\medskip

\begin{tcblisting}{listing only,breakable,
  colback=black!2,colframe=black!40,
  listing options={basicstyle=\ttfamily\footnotesize,breaklines=true}}
{
  "problem": "<the math problem here>",
  "chosen_pattern": "A or B",
  "chosen_solution": {
     "reasoning": "<one continuous paragraph>",
     "code_blocks": ["<one python block only>"],
     "outputs": ["<stdout lines>"],
     "final_answer": "..."
  },
  "counter_solution": {
     "reasoning": "<one continuous paragraph>",
     "code_blocks": ["<one python block only>"],
     "outputs": ["<stdout lines>"],
     "final_answer": "..."
  }
}
\end{tcblisting}
}
\end{tcolorbox}

\section{Additional Related Works on LLMS Reasoning Capacity}
Among various LLM abilities, foundational research in semantic matching~\cite{xue2023dual, xue2025structcoh} and temporal QA~\cite{xue2024qa} has paved the way for complex understanding. While domain adaptation~\cite{huang2023dtbs} and test-time integration~\cite{huang2025cosmic} enhance robustness, reasoning ability achieved by post-training~\cite{huang2026rlvr, lan2025mappo} stands as a key frontier. Given that traditional SFT may suffer from incomplete learning~\cite{xue2026sft}, recent works introduce uncertainty-aware reward modeling~\cite{xue2026reward}, dense feedback from LM critics~\cite{cao2024enhancing}, and preference-driven code generation~\cite{li2025preference} to refine alignment. Moreover, recent studies have advanced role-playing reasoning through human-actor emulation~\cite{chen2026actormindemulatinghumanactor} and improved continual learning via prototype-conditioned generative replay~\cite{chen-zeng-2025-prototype}.

The scope of reasoning has expanded into multimodal domains, where models now "think with 3D"~\cite{chen2025think} and utilize reinforced visual perception~\cite{chen2025visrl, chen2025sifthinker}. Advanced frameworks like OmniVideo-R1~\cite{chen2026omnivideo} and DV-Matcher~\cite{chen2025dv} further integrate audio-visual and geometric cues. Such multimodal intelligence significantly boosts retrieval performance, including chat-driven search~\cite{xie2025chat, xie2026conquer} and vision-driven video representation~\cite{xie2026hvd, xie2026delving}. Furthermore, safeguarding techniques~\cite{yu2024robust, lan2025contextual} and hallucination mitigation~\cite{yu2024mechanistic, ji2025calibrating} ensure the reliability of these complex systems.

To support deployment, system efficiency and continual learning have become indispensable. Inference is optimized via multi-tier KV cache management~\cite{chu2025mcam}, selective sharing~\cite{chu2025selective}, and adaptive MoE scheduling~\cite{chu2025dynamic, shen2025expertflow}. Moreover, lightweight techniques like parameter-efficient sampling~\cite{yao2024swift}, structured pruning~\cite{li2026sepprune}, and distillation~\cite{li2025frequency} enable efficient task execution, from medical imaging~\cite{Li2025Efficient} to web agents~\cite{ding2026dynaweb}. Finally, reversible alignment~\cite{xiao2026reversible} and unified LoRA generation~\cite{xiaometa, chen2024three} allow models to learn continually, while comprehensive interaction surveys~\cite{li2026comprehensive} and robust federated learning~\cite{chen2024confusion} continue to push the boundaries of scalable AI.

Beyond core architecture, recent research emphasizes a data-centric perspective on the entire LLM lifecycle~\cite{rao2025data}, where mathematical reasoning is further enhanced by adversarial data synthesis~\cite{yu2026mathagent} and iterative DPO with dynamic sampling~\cite{rao2025dynamic}. To improve model robustness and learning efficiency, advanced techniques such as difficulty-based sample weighting~\cite{zhou2022understanding}, implicit data augmentation~\cite{zhou2024boosting}, and adversarial training with anti-adversaries~\cite{zhou2024adversarial} have been introduced.

In the context of embodied AI and spatio-temporal reasoning, unified token spaces for human-object interaction~\cite{yang2026unihoi} and object-centric decoupling for robotic manipulation~\cite{yang2026instrucrobo} have improved explainable decision-making. These abilities are further integrated into 3D scene understanding~\cite{yang2026unibvr} and autonomous driving through spatio-temporal Chain-of-Thought (CoT)~\cite{zeng2025fsdrive} and dual-memory vision-language navigation~\cite{zeng2025janusvln}. Furthermore, the application of multi-scale graph learning and Transformers has shown significant potential in predicting complex real-world dynamics, including financial fraud detection~\cite{lei2026multiscale}, traffic flow prediction~\cite{shen2026mftformer}, disaster risk assessment~\cite{shen2025aienhanced}, mental health~\cite{yang2026exploring}, recommendations~\cite{wang2025improving} and socio-economic metrics such as labor market gradients~\cite{liu2026health} and business resilience indices~\cite{Shen2026bri}.
\section{Evaluate Pattern Chosen}
To further examine the reliability of Gemini’s pattern selection, we asked GPT-5 to independently select patterns without prior exposure to Gemini’s choices. Across 100 test cases, GPT-5’s selections were consistent with Gemini’s in 98 instances, corresponding to a 98\% agreement rate. This high level of concordance provides strong empirical support for the robustness of our approach. This follows LLM as a Judge~\cite{li2025generation,gao2023human,jiang2024memorization} method.

\end{document}